\documentclass{llncs}
\usepackage{makeidx}  

\usepackage{amsmath} 
\usepackage{amsfonts}

\usepackage{graphicx}
\usepackage{subfigure}
\usepackage{verbatim}

\begin{document}
%
%
\title{Extrinsic camera calibration method and its performance evaluation}
\titlerunning{Extrinsic camera calibration}  
%
\author{Jacek Komorowski\inst{1} \and Przemyslaw Rokita\inst{2}}
\authorrunning{J. Komorowski, P. Rokita}   
%
\tocauthor{Jacek Komorowski, Przemyslaw Rokita}
\institute{Maria Curie Sklodowska University\\
 Lublin, Poland\\
\email{jacek.komorowski@gmail.com}
\and
Warsaw University of Technology,
Institute of Computer Science,\\
Warszawa, Poland\\
\email{p.rokita@ii.pw.edu.pl}
}

\maketitle              

\begin{abstract}
This paper presents a method for extrinsic camera calibration (estimation of camera rotation and translation matrices) from a sequence of images. It is assumed camera intrinsic matrix and distortion coefficients are known and fixed during the entire sequence. 
Performance of the presented method is evaluated on a number of multi-view stereo test datasets.
Presented algorithm can be used as a first stage in a dense stereo reconstruction system.
\end{abstract}

\section{Introduction}
\label{jk-pr:sec:intro}

Motivation for development of the method described in this paper was our prior research on human face reconstruction from a sequence of images from a monocular camera.
Classical multi-view stereo reconstruction algorithms (as surveyed in \cite{jk-pr-calib:sei06}) assume fully calibrated setup, where both intrinsic and extrinsic camera parameters are known for each frame. 
Such algorithms cannot be used when an object moves or rotates freely in front of the camera or if a camera is moved to different positions (in uncontrolled manner) and a number of images of the static object is taken.
Even if camera intrinsic parameters are known and fixed during the entire sequence, extrinsic parameters (rotation matrix and translation vector relating camera reference frame with the world reference frame) are not known.
So before a multi-view stereo reconstruction algorithm can be used, a prior step to estimate camera pose for each image in the sequence is required.

Our approach is based on ideas used in modern structure from motion products such as Bundler \cite{jk-pr-calib:snav07} or Microsoft PhotoSynth.
These solutions work by finding geometric relationship (encoded by a fundamental matrix) between 2 images of the scene taken from different viewpoints. 
This is usually done by running a robust parameter estimation method (e.g. RANSAC) combined with 7-point or 8-point fundamental matrix estimation algorithm using putative pairs of corresponding features from 2 images.

In contrast to aforementioned solutions, we assume fixed and known camera intrinsic parameters. 
Making such assumptions is beneficial for 2 reasons. First, less pairs of corresponding points are required to recover 2-view scene geometry. Second, fundamental matrix estimation algorithms do not work for planar surfaces (so called planar degeneracy) \cite{jk-pr-calib:har04}. When intrinsic matrix is known essential matrix can be estimated instead of fundamental matrix.
So assuming known intrinsic parameters is advantageous when processing sequences of images of low texture (e.g. depicting a rotating human face) or planar objects.

Unfortunately known algorithms for essential matrix estimation from 5 pairs of corresponding points are very complex and implementations is not freely available.
E.g. Nister 5-point algorithm \cite{jk-pr-calib:nis04} requires SVD, partial Gauss-Jordan elimination with pivoting of a system of of polynomial equations of the third degree and finally finding roots of a 10th degree polynomial. Such complexity can potentially lead to significant numerical errors and make such methods inapplicable in practice.

The aim of this paper is twofold: first to present a solution for estimation of extrinsic parameters from a sequence of images taken by a calibrated camera, and second, to asses the accuracy of the presented method on various datasets.

\section{Extrinsic camera calibration method details}
\label{jk-pr:sec:method}

An input to our extrinsic camera calibration method is a sequence of images from a monocular camera, such as depicted on Fig. \ref{jk-pr:fig10}. It is assumed camera intrinsic matrix $\mathbf{K}$ and distortion coefficients are known and fixed for the entire sequence. The following steps are done to recover camera extrinsic parameters for each image in the input sequence:

\begin{enumerate}
	\item Initial processing: geometric distortions removal and (when required) object segmentation from the background. In scenarios where camera is moving (such as a sequence depicted on Fig. \ref{jk-pr:fig10:2}) segmentation is not necessary. When camera is fixed and an object is moving (e.g. placed on a rotating turnstile, such as Fig. \ref{jk-pr:fig10:1}), the object should be segmented from the background. Further processing is done on undistorted and segmented images. 
	\item Detection of SIFT features on all images in the sequence. 
	\item Estimation of the relative pose between 2 initial images in the sequence:
	\begin{enumerate}
		\item Finding pairs of putative matches between SIFT features on both images.
		\item Computation of essential matrix $E_{12}$ relating two images using RANSAC \cite{jk-pr-calib:fis81} with Nister \cite{jk-pr-calib:nis04} solution to 5-point relative pose problem. The relative pose (translation vector $T_{2}$ and rotation matrix $R_{2}$) is recovered from $E_{12}$ as described in \cite{jk-pr-calib:nis04}.
		\item Construction of a sparse 3D model (as a point cloud $\left\{ X_i \right\}$) by metric triangulation of pairs of consistent features from two images. 
		\item 3D points and camera pose refinement using bundle adjustment method \cite{jk-pr-calib:tri99} to minimize reprojection error.
	\end{enumerate}
	
	\item Iterative estimation of an absolute pose of each subsequent image $I_n$ with respect to 3D model built so far:
	\begin{enumerate}
		\item Finding putative matches between features on the image $I_n$ and 3D points already in the model.
		\item Computation of an absolute pose (translation vector $T_{k}$ and rotation matrix $R_{k}$) of the image $I_k$ with respect to the 3D model. This is done using RANSAC \cite{jk-pr-calib:fis81} with Finsterwalder 3-point perspective pose estimation algorithm \cite{jk-pr-calib:har94}.
		\item Guided matching of features from currently processed image $I_k$ and images processed in the previous steps. New 3D points are generated and added to 3D model (and support of existing 3D points is extended)  by metric triangulation of matching features.
		\item 3D points and camera pose refinement using bundle adjustment method \cite{jk-pr-calib:tri99} to minimize reprojection error.
		\item Removal of 3D points with the worst support
	\end{enumerate}
	
\end{enumerate}

Additional notes on algorithm steps:

\begin{figure}
\centering
\subfigure[]{
\includegraphics[height=2.4 cm]{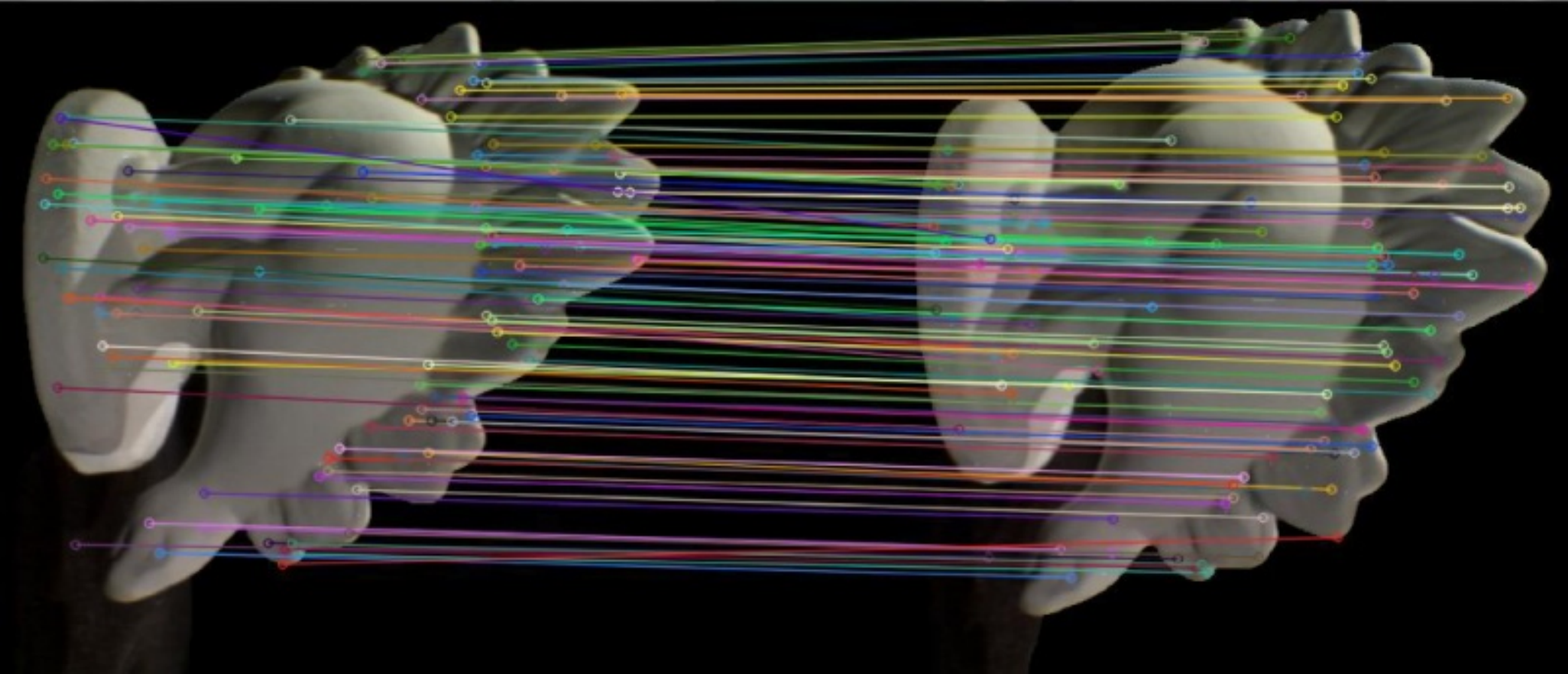}
\label{jk:fig:200}
}
\subfigure[]{
\includegraphics[height=2.4 cm]{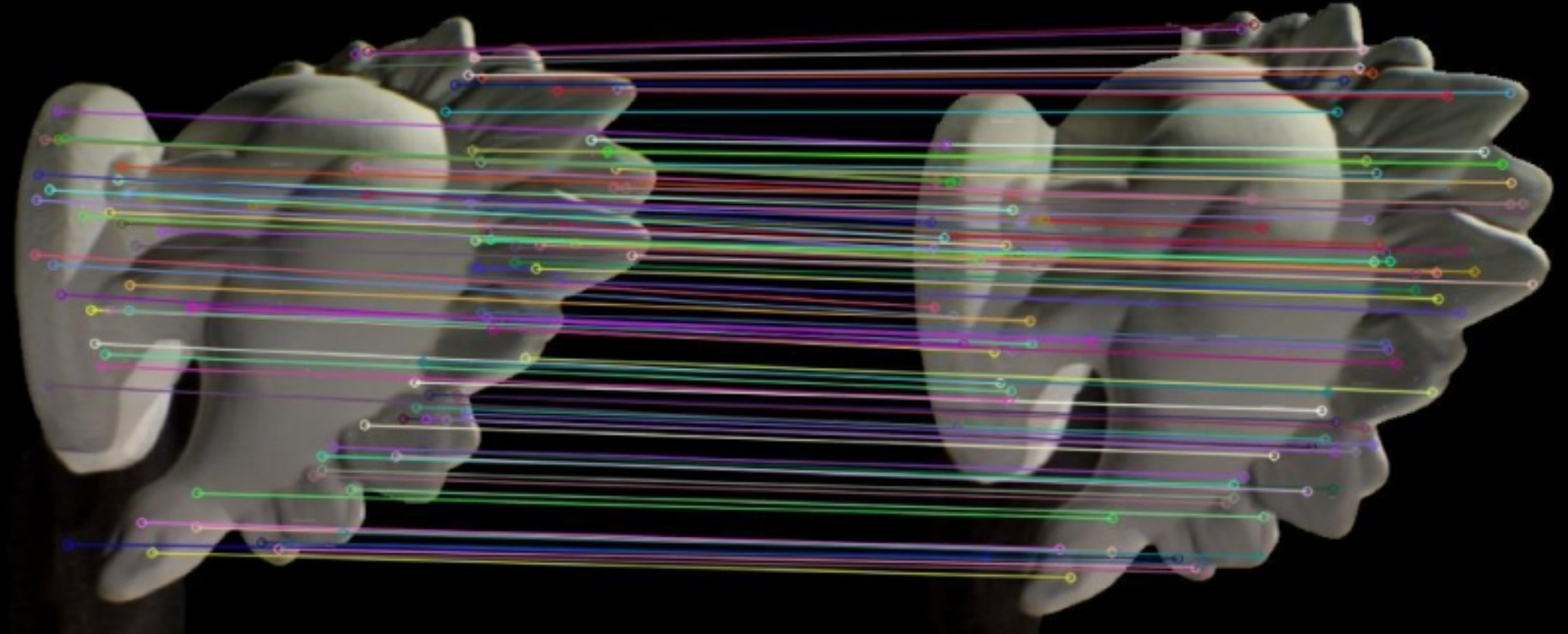}
\label{jk:fig:201}
}

\label{jk:fig:202}
\caption{Pairs of matches between 2 images \subref{jk:fig:200} putative matches \subref{jk:fig:201} matches consistent with epipolar geometry encoded by estimated essential matrix $E$} 
\end{figure}

\paragraph{Step 2} 
SIFT features \cite{jk-pr-calib:lowe99} are a common choice in modern structure from motion solutions. This is dictated by their invariance to scaling, rotation and, to some extent, lighting variance and small affine image transformations. These properties are important when finding corresponding features on images taken from different viewpoints. At this step SIFT features are found and feature descriptors (represented as vectors from $\mathbb{R}^{128}$) are computed for each image in the sequence. 
Threshold of SIFT feature detector is dynamically tuned to ensure there's a sufficient number of keypoints found on each image. In our implementation if there's less than 500 keypoints found the detector is re-run with decreased threshold.

As camera intrinsic matrix $\mathbf{K}$ is known and fixed during entire sequence, coordinates of detected features are \emph{normalized}, that is multiplied by $\mathbf{K}^{-1}$. This  is equivalent to assuming that camera intrinsic matrix is identity. All further processing is done using \emph{normalized} coordinates and assuming camera intrinsic matrix is identity.

\paragraph{Step 3a} For each keypoint from the first image the closest (in the feature descriptor space) keypoint from the second image is found. Only pairs fulfilling nearest neighbour ratio criterion (that is ratio of a distance to the corresponding keypoint to the distance to the second-closest keypoint on the other image is below given threshold $\Theta = 1.25$) are kept as putative matches. See Fig. \ref{jk:fig:200}.

\paragraph{Step 3b} RANSAC \cite{jk-pr-calib:fis81} robust parameter estimation is used with our implementation of Nister 5-point algorithm \cite{jk-pr-calib:nis04} to estimate relative pose between 2 cameras. 
In each RANSAC iteration 5 pairs (the minimum number of correspondences needed to estimate relative pose between 2 cameras) of potentially matching keypoints are sampled at random from a set of putative correspondences and an essential matrix $E$ is estimated using the chosen sample. 
The estimation that produces the biggest number of inliers (that is putative matches consistent with epipolar geometry induced by an estimated essential matrix $E$) is kept.
Results of this step are: essential matrix $E_{12}$ describing stereo geometry between first 2 images, rotation matrix $\mathbf{R}_2$ and translation vector $T_2$ describing the relative pose of the second image with respect to the first image and RANSAC consensus set consisting of pairs of matching features consistent with epipolar geometry (see Fig. \ref{jk:fig:201})

\paragraph{Step 3c} 
3D model is constructed by metric triangulation of pairs of compatible features from two images. Two criteria are taken into account: \emph{visual compatibility} (Euclidean distance between descriptors of corresponding features is below a threshold) and \emph{geometric compatibility} (reprojection error is below a threshold).
First all points from RANSAC consensus set are used to construct 3D points by metric triangulation.
Then additional matches between 2 images are sought with a guided matching.

\paragraph{Step 3d} 
Bundle adjustment method \cite{jk-pr-calib:tri99} minimizes total reprojection error by joint optimization of camera poses and 3D points position using Levenberg-Marquardt nonlinear optimization algorithm.
\footnote{sba sparse bundle adjustment library \cite{jk-pr-calib:lour09} available at
\url{http://www.ics.forth.gr/~lourakis/sba/}
is used in our implementation
}
It is assumed world coordinate frame aligns with the first camera coordinate frame, so first camera pose is fixed $\left( R_1= \mathbf{I}, T_1= \mathbf{0} \right)$ and only second camera pose and 3D points coordinates are optimized by minimizing:
\begin{equation} \label{jk:eq:ba}
\min_{
R_2, T_2,
\left\{
X_i
\right\}
}
\sum_{i=1}^{2} 
\sum_{j=1}^{N} 
\left\|
R_i \left( X_j - T_i \right)
-
x_j^i
\right\|
\end{equation}
where
$x_j^i$ are coordinates of the feature from $i$-th image used to construct 3D point $X_j$ and
$R_i \left( X_j - T_i \right)$ is a projection of a 3D point $X_j$ onto $i$-th image.

\paragraph{Step 4c} 
For each keypoint  $k$ from a currently processed image $I_n$ a number (20 in our implementation) of closest (in feature descriptor space) keypoints from already processed, nearby images is sought. Features visually and geometrically compatible with $k$ are retained.
If $k$ has compatible features from at least 2 other images a new 3D point is constructed by metric triangulation and compatible keypoints form its support.

\paragraph{Step 4e} 
Support of all 3D points is verified and features which are not geometrically compatible (e.g. due to cameras pose or 3D points position refinement) are removed.
Then 3D points not having a support from at least 3 images are removed.

\begin{figure}
\centering
\includegraphics[height=4 cm]{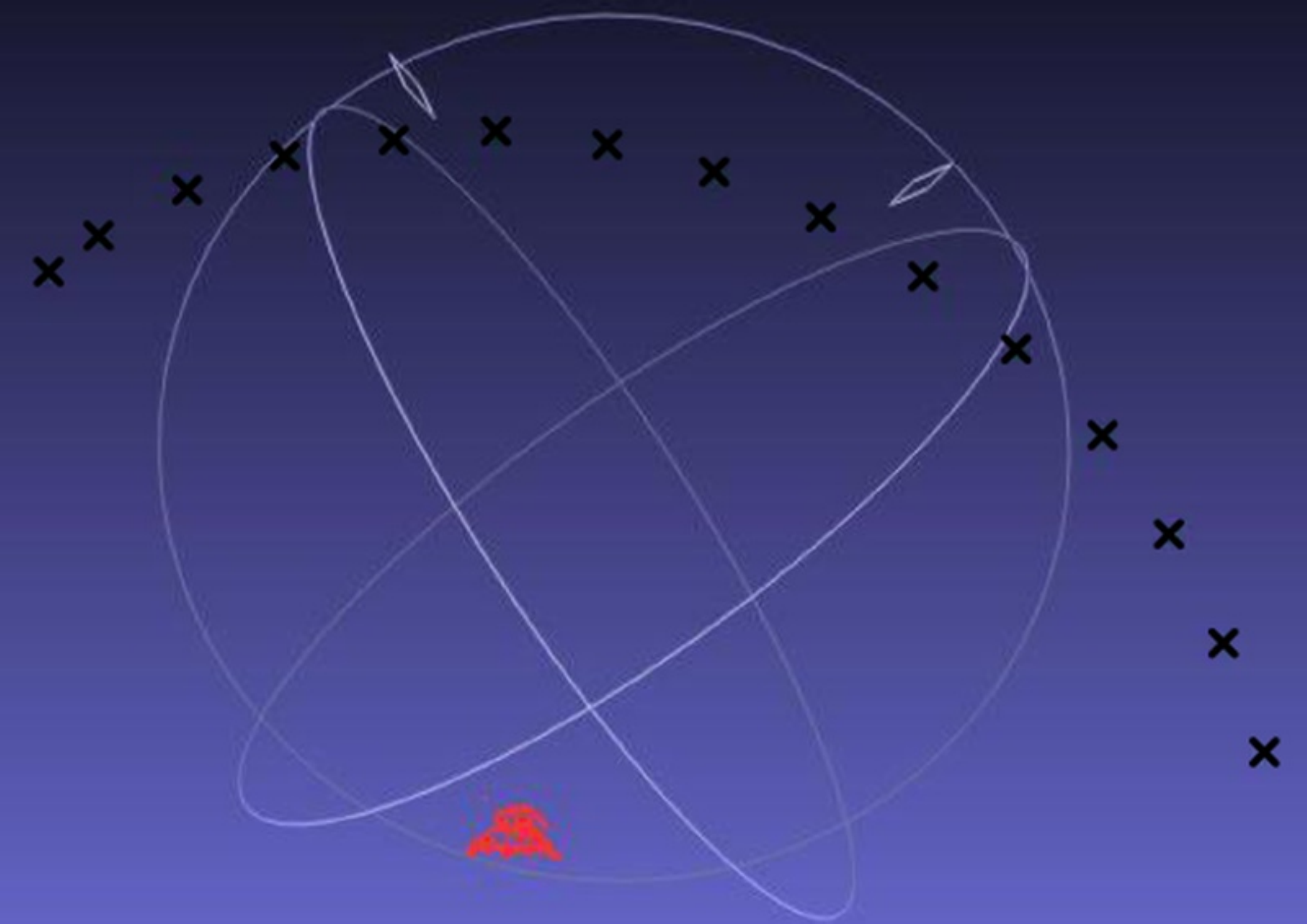}
\caption[]{Estimated camera poses (black crosses) and a sparse object model (dots) recovered from a Dino dataset (Fig. \ref{jk-pr:fig10:1}}
\label{jk-pr:fig:100}
\end{figure}

Final results are depicted on Fig. \ref{jk-pr:fig:100}, where black crosses represent recovered camera poses for Dino input sequence (Fig. \ref{jk-pr:fig10:1}).

\section{Experiments}

\begin{figure}
\centering

\subfigure[Dino]{\includegraphics[height=2.5 cm]{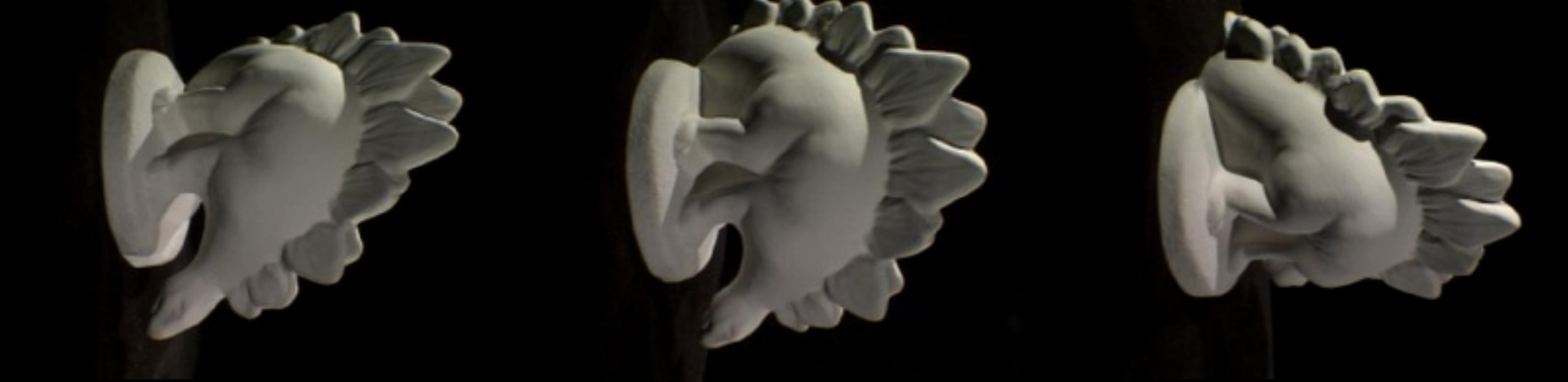}
\label{jk-pr:fig10:1}
}

\subfigure[Fountain-P11]{\includegraphics[height=2.5 cm]{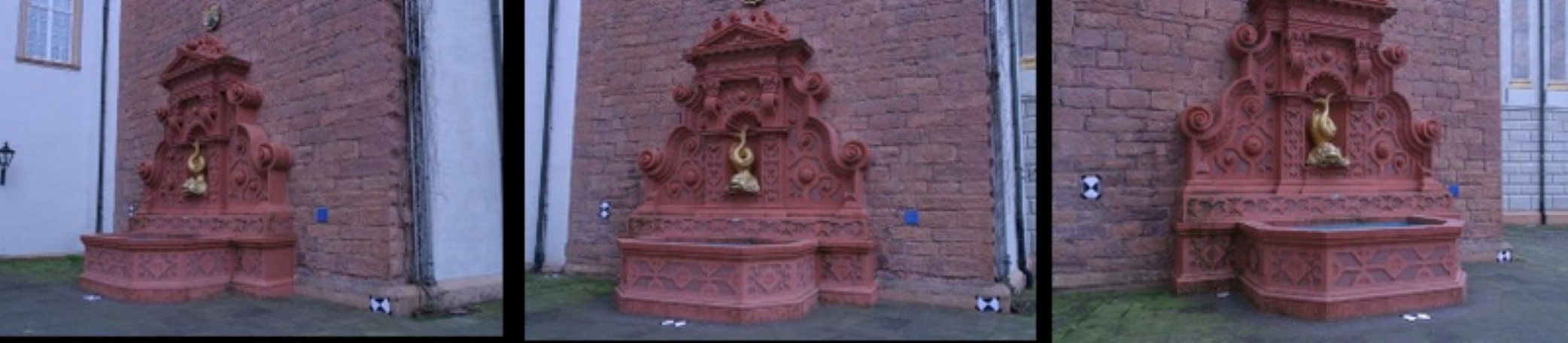}
\label{jk-pr:fig10:2}
}

\subfigure[Castle-P19]{\includegraphics[height=2.5 cm]{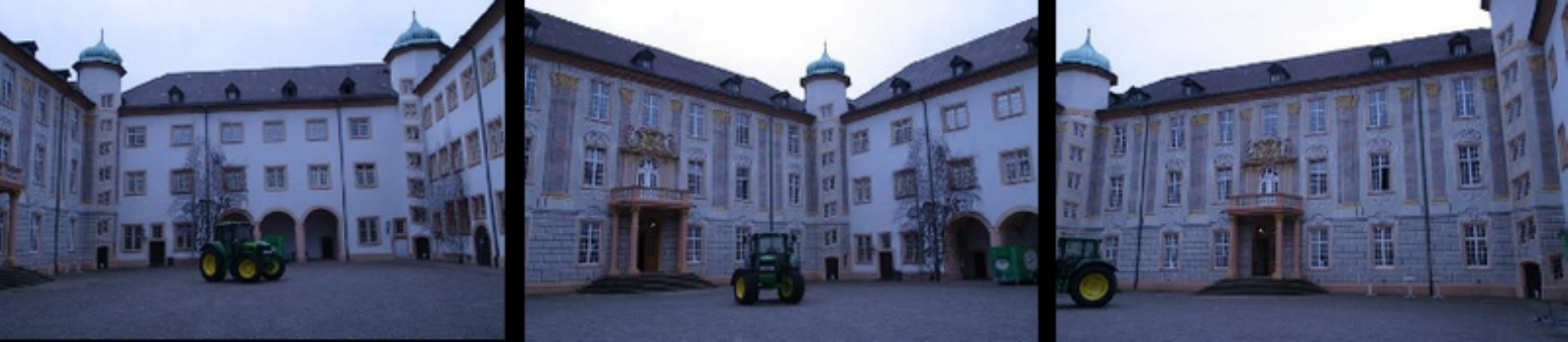}
\label{jk-pr:fig10:3}
}

\subfigure[Entry-P10]{\includegraphics[height=2.5 cm]{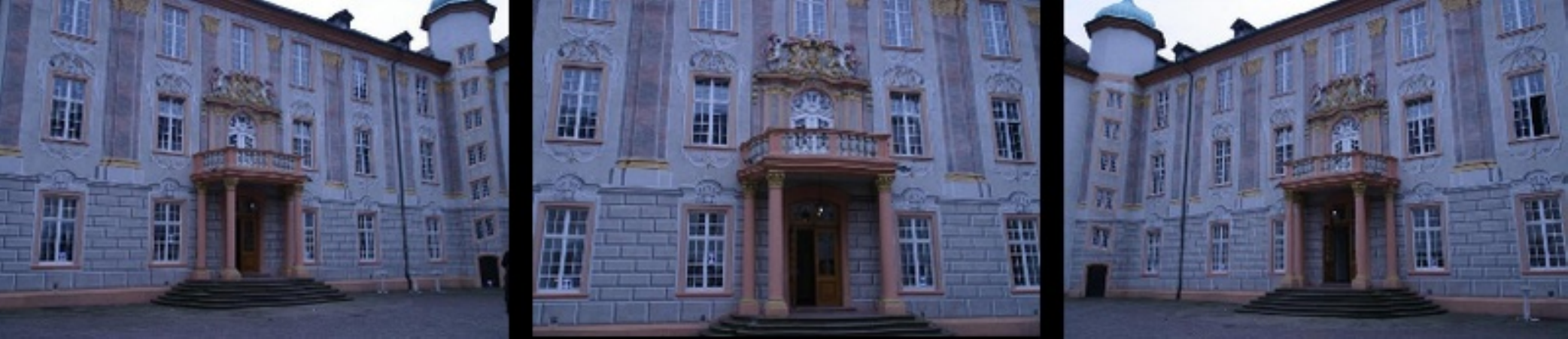}
\label{jk-pr:fig10:4}
}

\subfigure[Herz-Jesu-P8]{\includegraphics[height=2.5 cm]{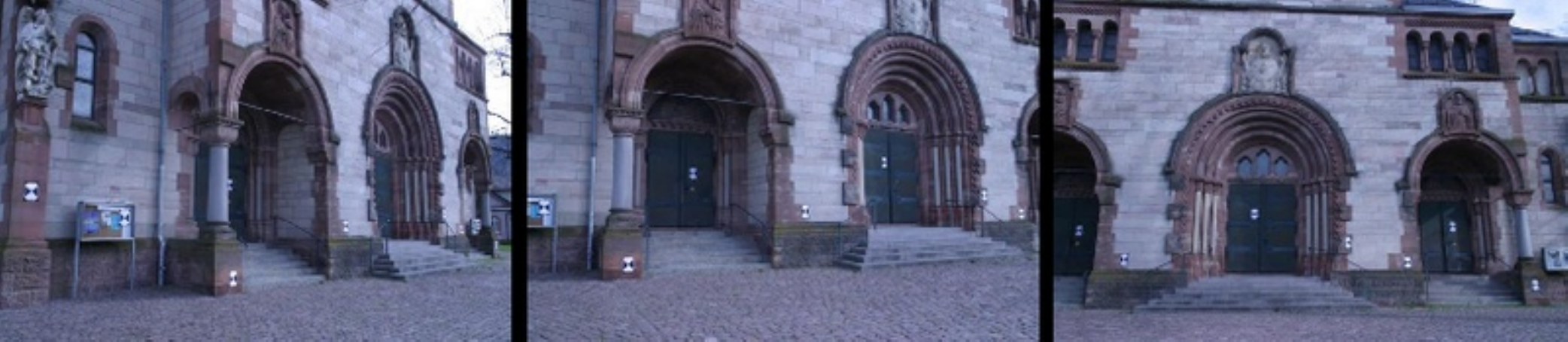}
\label{jk-pr:fig10:5}
}

\caption[]{Exemplary images from datasets used in experiments.}
\label{jk-pr:fig10}

\end{figure}

An accuracy of the proposed method was evaluated quantitatively using five multi-view stereo datasets with given camera intrinsic and extrinsic parameters.
The following multi-view stereo datasets were used:
\begin{itemize}
	
	\item Dataset \cite{jk-pr-calib:sei06} 
	\footnote{Available at \url{http://vision.middlebury.edu/mview/data/}} 
	(Fig. \ref{jk-pr:fig10:1}) 
	containing sequences of images (640x480 pixels) of a plaster dinosaur sampled every 7.5 degree on a ring around it.
	This seems a very demanding dataset for automatic recovery of camera pose as an object is almost textureless and relatively few distinctive keypoints can be found on each image.
	
	\item Dataset \cite{jk-pr-calib:stre08} \footnote{Available at \url{http://cvlab.epfl.ch/~Strecha/multiview/denseMVS.html}} 
	(Fig. \ref{jk-pr:fig10:2}, \ref{jk-pr:fig10:3}, \ref{jk-pr:fig10:4}, \ref{jk-pr:fig10:5}) containing  sequences of high resolution (3072x2048 pixels) images of architectural objects. Number of images in each sequence vary from 8 to 19. 
 
\end{itemize}

\begin{figure}
\centering

\subfigure[{Dino dataset (Fig \ref{jk-pr:fig10:1}})]{
\includegraphics[width=8cm]{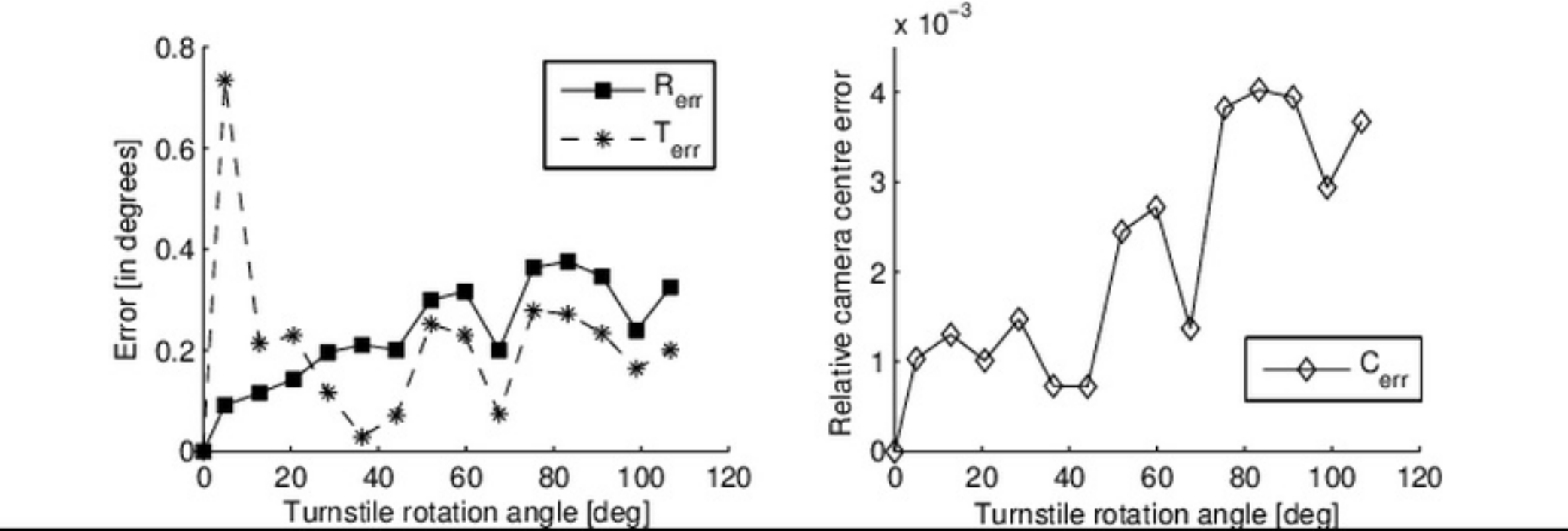}
\label{jk:fig:51}
}

\subfigure[{Fountain-P11 dataset (Fig \ref{jk-pr:fig10:2}})]{
\includegraphics[width=8cm]{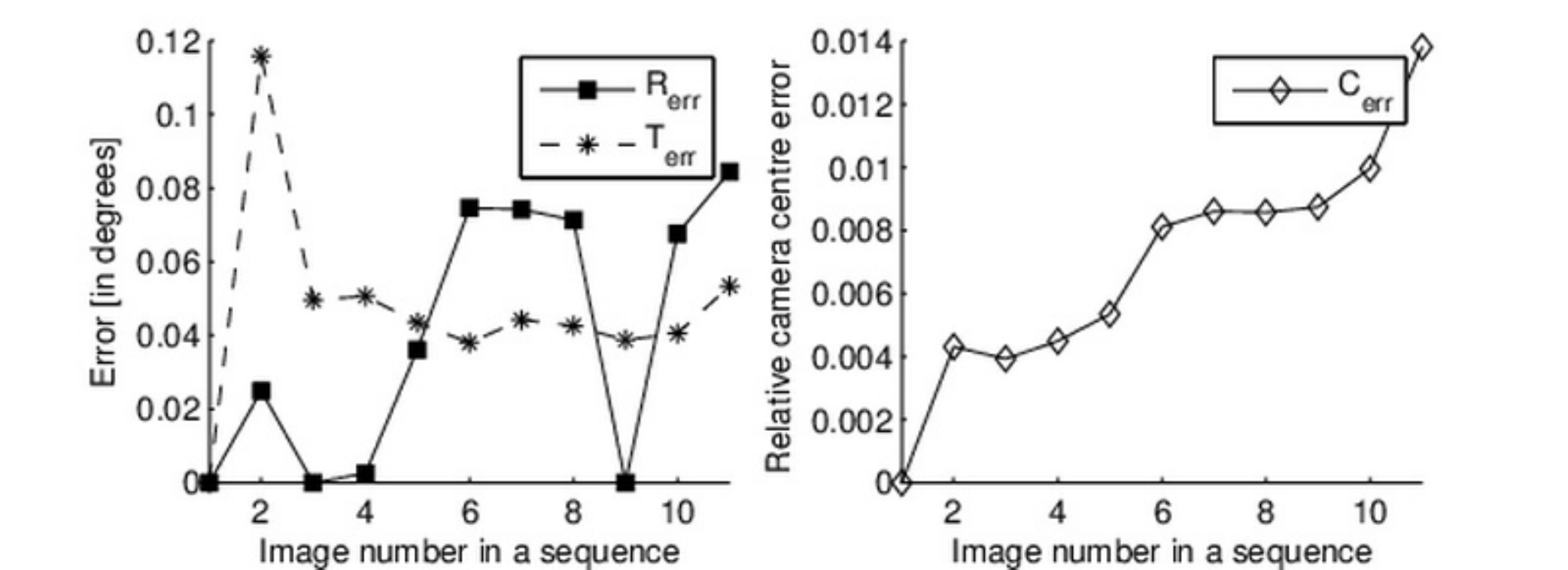}
\label{jk:fig:52}
}

\subfigure[{Castle-P19 dataset (Fig \ref{jk-pr:fig10:3}})]{
\includegraphics[width=8cm]{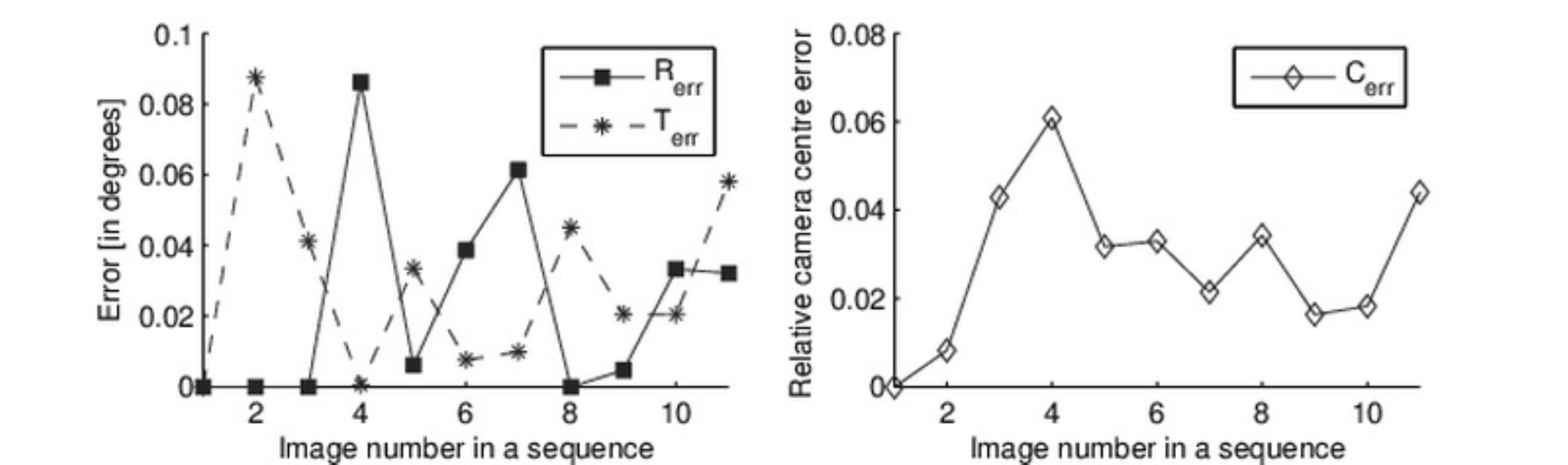}
\label{jk:fig:53}
}

\subfigure[{Entry-P10 dataset (Fig \ref{jk-pr:fig10:4}})]{
\includegraphics[width=8cm]{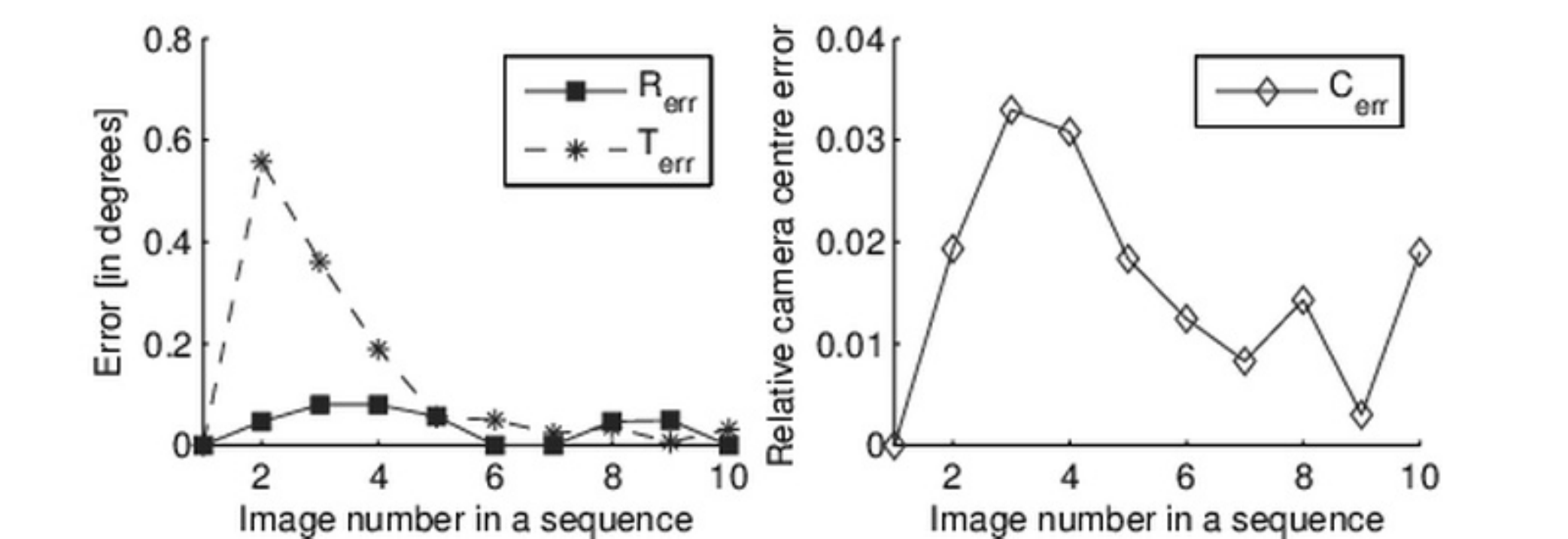}
\label{jk:fig:54}
}

\subfigure[{Herz-Jesu-P8 dataset (Fig \ref{jk-pr:fig10:5}})]{
\includegraphics[width=8cm]{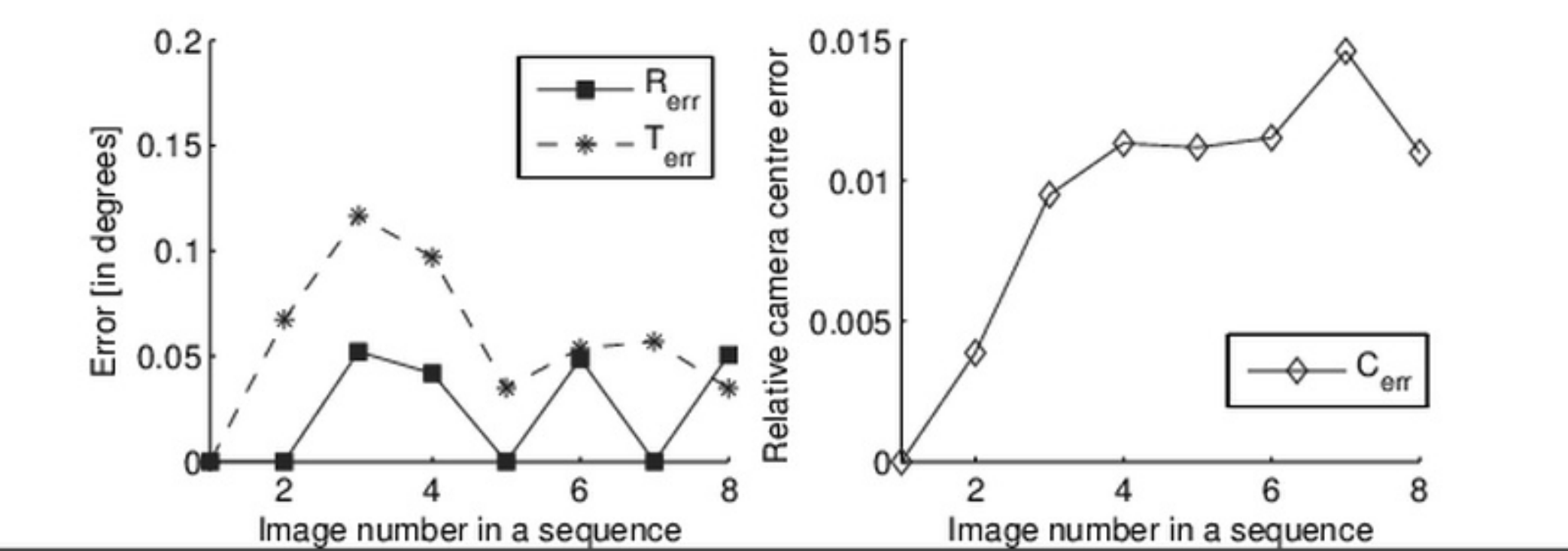}
\label{jk:fig:55}
}

\label{jk:fig:50}
\caption{Performance of extrinsic camera calibration method on test datasets. Left column: rotation error $R_{err}$ and translation error (angular component) $T_{err}$, right column: relative camera center error $C_{err}$} 
\end{figure}

\paragraph{Error measures} 
Rotation error $R_{err}$ is measured as the rotation angle needed to align ground truth rotation matrix $R_i$ and estimated rotation matrix $\hat{R}_i$ for $i$-th image in the sequence.
$$
R_{err} = \cos^{-1} \frac{\mathrm{Tr}\left(\Delta R_i \right)-1}{2},
$$
where $\Delta R_i = R_i^{-1} \hat{R_i}$ is the rotation matrix that aligns estimated rotation $\hat{R}_i$ with the ground truth rotation $R_i$ and $\mathrm{Tr} ( \Delta R_i )$ is a trace of $\Delta R_i$.

It is not possible to directly compare a ground truth translation vector $T_i$ and estimated translation vector $\hat{T}_i$, as estimation of camera extrinsic parameters given known intrinsic parameters is possible only up to a scale factor. 
Angular component of translation error, that is an angle between true translation vector $T_i$ and estimated translation vector $\hat{T}_i$, is examined using the formula:
$$T_{err} = \cos^{-1} \left( \frac{\hat{T}_i \cdot T_i}{| \hat{T}_i | |T_i| }\right)$$

In order to verify accuracy of translation vector estimation the following procedure is deployed.
First the scale $s$ between two point clouds consisting of ground truth and estimated camera centres is calculated using the formula taken from \cite{jk-pr-calib:hor87}:
$$
s = 
\sqrt{
\frac{
\sum_{i=1}^{n}
\left\| 
T_i - T'
\right\| ^ 2
}
{
\sum_{i=1}^{n}
\left\| 
\hat{T}_i - \hat{T}'
\right\| ^ 2
}
} \ \ ,
$$
where $T' = \frac{1}{n} \sum_{i=1}^n T_i$ is a centroid of a point cloud consisting of ground truth camera centres 
and $\hat{T}' = \frac{1}{n} \sum_{i=1}^n \hat{T}_i$ is a centroid of estimated camera centres cloud.
Then estimated translation vectors are brought to the same scale with the ground truth by multiplying by $s$.
Unfortunately ground truth coordinate system scale is different in different datasets and often not given in physical units. To allow a meaningful interpretation an estimated error is renormalized using a distance between first a second ground truth camera centres as a unit. This gives the final formula for relative camera centre error:

$$
C_{err} = \frac{1}{\alpha} \left( s * \hat{T}_i - T_i \right) \ \ ,
$$
where $\alpha = T_0 - T_1$ is a difference between first and second ground truth camera centres.


\paragraph{Results}
Error metrics of our extrinsic camera calibration method on test datasets are depicted on Fig. \ref{jk:fig:50}.
Results for outdoor sequences (Fountain-P11, Castle-P19, Entry-P10, Herz-Jesu-P8) 
are comparable.
Rotational error $R_{err}$ is small and varies between 0 and 0.1 degree.
Angular component of translational error $T_{err}$ is also small and varies between 0 and 0.1 degree with the exception of Entry-P10 dataset, where it peaks above 0.5 for second image in the sequence. 
This is likely caused by an incorrect estimation of the relative pose of the second camera with respect to the first camera. Images in Entry-P10 dataset contain many repetitive structures (windows) and some incorrect matches must have been established between keypoints from first two images. 

Relative camera center error $C_{err}$ shows two different characteristics.
For Dino, Fountain-P11, Herz-Jesu-P8 datasets it increases for further images in the sequence,
whereas for Castle-P19 and Entry-P10 no trend can be noticed and the error doesn't grow.
Castle-P19 and Entry-P10 datasets contain a lot of highly distinctive keypoints (e.g. window corners) which are matched across multiple distant images.
In each case relative camera centre error $C_{err}$ is below 0.06 which means that
if distance between first and second camera is 1m then each camera is positioned with 6 cm accuracy. 
In all test datasets distance between cameras associated with the first and last image in the sequence is about 10 times bigger than the distance between 2 first cameras. So distant cameras, 10 m apart, are localized with 6 cm accuracy.

Algorithm also performs quite well for the most demanding dataset: Dino.
Even for distant frames (where object was rotated over 100 degrees from its initial position) rotation error $R_{err}$ is below 0.4 degree.

\section{Conclusions and Future Work}

Conducted experiments proved that the presented method can be used to recover camera extrinsic parameters from a sequence of images quite accurately.
On all but one datasets relative camera center error
$C_{err}$ was below 0.06 which means that 
if distance between first and second camera is 1m 
all other cameras are located within 6 cm accuracy.
For some sequences results are significantly better, e.g. for Fountain-P11 and Herz-Jesu-P8
$C_{err}$ is below 0.015 which means cameras are located with 1.5 cm accuracy (assuming 1 m distance between first and second camera).

The algorithm performed very well on an objects with relatively little texture (Dino sequence).
Errors in subsequent images were increasing (as there were very few keypoints visible from a wide angle)
but even when an object rotated almost 120 degrees rotation matrix estimation error $R_{err}$ was below 0.4 degree.

In the future it is planned to use the presented method as a first stage in a dense stereo reconstruction system.
After camera pose is estimated for each image in the sequence multi-view stereo reconstruction method will be used to generate a dense point cloud representing an object.

%
%


\begin{thebibliography}{10}
%

\bibitem{jk-pr-calib:fis81} 
Fischler M, Bolles R:
Random Sample Consensus: A Paradigm for Model Fitting with Applications to Image Analysis and Automated Cartography. Communications of the ACM (1981)

\bibitem{jk-pr-calib:har04} 
Hartley R, Zisserman A:
Multiple View Geometry in Computer Vision.
Cambridge University Press (2004)

\bibitem{jk-pr-calib:har94} 
Haralick R, Lee C, Ottenberg K, Nolle M:
Review and analysis of solutions of the three point perspective pose estimation problem.
International Journal of Computer Vision (1994)

\bibitem{jk-pr-calib:hor87} 
Horn B:
Closed-form solution of absolute orientation using unit quaternions.
Journal of the Optical Society of America, Vol. 4 (1987)

\bibitem{jk-pr-calib:nis04} 
Nister D:
An efficient solution to the five-point relative pose problem.
IEEE Transactions on Pattern Analysis and Machine Intelligence (2004)

\bibitem{jk-pr-calib:lour09} 
Lourakis M, Argyros A:
SBA: A Software Package for Generic Sparse Bundle Adjustment.
ACM Trans. Math. Software, vol. 36 (2009)

\bibitem{jk-pr-calib:lowe99} 
Lowe, D:
Object recognition from local scale-invariant features.
Proceedings of the International Conference on Computer Vision (1999)

\bibitem{jk-pr-calib:snav07} 
Snavely, N et al.:
Modelling the World from Internet Photo Collections
International Journal of Computer Vision (2007)

\bibitem{jk-pr-calib:tri99} 
Triggs B, McLauchlan P, Hartley R, Fitzgibbon A:
Bundle Adjustment — A Modern Synthesis.
Proceedings of the IWVA (1999)

\bibitem{jk-pr-calib:sei06} Seitz, S. et al.:
A Comparison and Evaluation of Multi-View Stereo Reconstruction Algorithms.
CVPR 2006, Vol. 1 (2006)

\bibitem{jk-pr-calib:stre08} 
Strecha C., von Hansen W., Van Gool L., Fua P, Thoennessen U.:
On Benchmarking Camera Calibration and Multi-View Stereo for High Resolution Imagery.
CVPR (2008)

\end{thebibliography}
\end{document}